# 1st Place Solution to Google Landmark Retrieval 2020

SeungKee Jeon(Samsung Electronics)*

123use321@gmail.com

## Abstract

*This paper presents the 1st place solution to the Google Landmark Retrieval 2020 Competition on Kaggle. The solution is based on metric learning to classify numerous landmark classes, and uses transfer learning with two train datasets, fine-tuning on bigger images, adjusting loss weight for cleaner samples, and esemble to enhance the model's performance further. Finally, it scored 0.38677 mAP@100 on the private leaderboard.*

## 1.Introduction

Google Landmark Retrieval 2020 Competition[1] is the third landmark retrieval competition on Kaggle. The task of image retrieval is to rank images in an index set by their relevance to a query image. In past landmark retrieval competitions, the developed models were expected to retrieve database images containing the same landmark as query.

To put the emphasis on representation learning, this competition requires you to create a model that extracts a feature embeddings from the images, then scoring system will use the model to 1) Extract embeddings for the private test and index sets 2) Create a kNN(k=100) lookup for each test sample, using the Euclidean distance between test and index embeddings 3) Score the quality of the lookups using the competition metric. The public test, index image sets are subsets of Google Landmarks Dataset v2(GLD2)[2], while private test, index image sets are completely new datasets.

GLD2 is the biggest landmark dataset, which contains images annotated with labels representing human-made and natural landmarks. It contains approximately 5 million images, split into 3 sets of images: train, index and test. There are 4132914 images in train set, 761757 images in index set, and 117577 images in test set.

GLD2 is constructed by mining web landmark images, so it is very noisy. There is a cleaned version of GLD2(CGLD2)[3], which was made by team smlyaka using automatic data cleaning system for Google Landmark Retrieval 2019 Competition[4].

The train set of GLD2 contains 4132914 images and 203094 classes, while the train set of CGLD2 contains 1580470 images and 81313 classes. Both GLD2 train set and CGLD2 train set are used for training in this solution.

In the rest, I describe the basic configurations in Section2, training strategies in Section3, ensemble method in Section4, and summary in Section5.

## 2.Basic Configuration

The model structure is depicted in Figure 1. Efficientnet[5] and global average pooling extract features from images, and a deep neural network is followed to squeeze the features into smaller dimensions for compact representation and reducing model size. After that, cosine softmax[6] is used to classify a number of classes. Imagenet pretrained efficientnet[7] was used for backbone CNNs at initial stage, and embedding feature size 512 was used for all models. For cosine softmax parameters, scale value was automatically determined by fixed adacos[8]. Margin value was set to 0, because train datasets are noisy so trying to cluster more between same class samples could make training more difficult. To deal with imbalanced classes, weighted cross entropy was used, and weight was determined proportional to 1/log(class count) for each class. For image augmentation, only left-right flip was used since there was low possibility of overfitting due to the large number of samples, and not to disturb the image distributions. Stochastic gradient descent optimizer was used for training, where learning rate, momentum, weight decay are set to 1e-3, 0.9, 1e-5. No learning scheduling was used for training. For validation set, 1 sample per class which has equal to or larger than 4 samples in CGLD2 was used, as a result 72322 samples were used for validation. Validation loss was calculated using non-weighted cross entropy. It is chosen this way because I wanted to have as much as classes in validation set, while giving them same importance. It is quite small size compared to large training data size, but validation loss correlated well with the leaderboard score. Google Colab[9] was used for all experiments, which provides TPUv2-8.

---

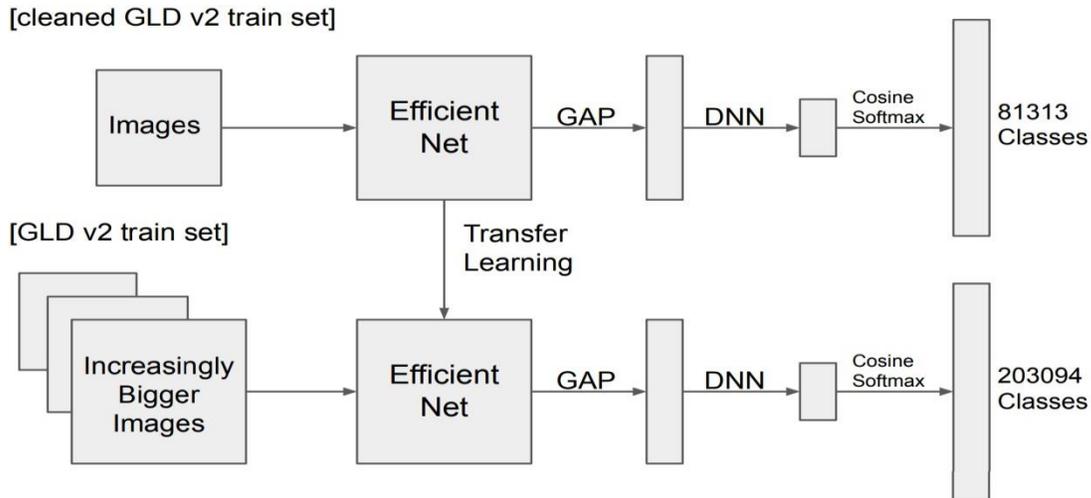

Figure 1. Basic model structure

## 3. Training Strategy

In the first step, CGLD2 was used to train the model to classify 81313 landmark classes. Single efficientnet7 backbone based model with 512x512 image inputs and batchsize 64 took 35 epochs, or 149 hours for validation loss to converge. It scored private LB score of 0.30264.

Second, GLD2 was used to train new model to classify 203094 classes, where efficientnet backbone is taken from step 1 for transfer learning. With batch size 64, it took 13 epochs, or 150 hours for validation loss to converge. It scored 0.33749, which is huge improvement from before. This process was the main driving force to increase the score. From here, it was found that even GLD2 is noisy, it helps to make image feature embeddings to be more representative. Given that training on GLD2 was successful, I tried to train the model from scratch using GLD2 for a long time, but validation loss slowly decreased and leaderboard score was much lower. This experiment showed that CGLD2 is actually cleaner than GLD2 and it helps the model to learn important CNN filters.

Third, whole model from step 2 was used as is and was given increasingly bigger images. 640x640 image inputs were given with batch size 64 to the model, and it took 5 epochs or 80 hours for validation loss to converge. Next, 736 x736 image inputs were given with batch size 32, and it took 3 epochs or 84 hours for the loss to converge. It was found that bigger the images, better the scores. This step scored 0.35389 on private LB for model with 640x640 inputs, and 0.36364 for model with 736x736 inputs. Since only input images changed and the whole model was reused, training was able to converge quite fast. And some data augmentation effect was also expected, because different image sizes could give CNN filters a new challenge to optimize.

Fourth, whole model from step 3 was taken and loss weight for CGLD2 samples were set twice. It was derived from the experience on step 2, that training too long with GLD2 could worsen the model performance. By changing the loss weights, I wanted the model to focus more on classifying CGLD2 samples. For model with 640x640 inputs from step3, it took 4 epochs, or 64 hours for validation loss to converge with batch size 64. For model with 736x736 inputs from step3, it took 3 epochs, or 84 hours for validation loss to converge with batch size 32. It scored 0.35932 for model with 640x640 image inputs, and 0.36569 for model with 736x736 image inputs, which showed that this step was effective.

|  | public score | private score |
|---|---|---|
| step1, 512x512 | 0.33907 | 0.30264 |
| step2, 512x512 | 0.36576 | 0.33749 |
| step3, 640x640 | 0.39121 | 0.35389 |
| step3, 736x736 | 0.40174 | 0.36364 |
| step4, 640x640 | 0.39881 | 0.35932 |
| step4, 736x736 | 0.40215 | 0.36569 |

Table 1. Leaderboard scores for each step. It represents the scores of single efficientnet7 backbone based model.

## 4. Ensemble

Ensemble method was used to raise the leaderboard score further. Feature embeddings from several models were concatenated with weight to make final feature embeddings. Used models are one efficientnet7, one efficientnet6, and two efficientnet5 backbone models. Weights were given based on the performance of each model. 1.0 for efficientnet7, 0.8 for efficientnet6, and 0.5 for efficientnet5 were finally chosen by inspecting few submission results.

With all models that went through step3, it scored 0.38366 on the private leaderboard. And by applying step4 for efficientnet7 backbone models, it scored 0.38677 on the private leaderboard, which is the best score I've got.

## 5. Summary

In this paper, 1st place solution for Google Landmark Retrieval 2020 was presented in detail. The solution used metric learning to classify numerous landmark classes, and gradually increased the leaderboard score by adopting transfer learning with two train datasets, finetuning on bigger images, adjusting loss weights for cleaner train samples, and finally ensemble method.

## References

[1] https://www.kaggle.com/c/landmark-retrieval-2020
[2] T. Weyand, A. Araujo, B. Cao, J. Sim. Google Landmarks Dataset v2 - A Large-Scale Benchmark for Instance-Level Recognition and Retrieval. CVPR, 2020
[3] Kohei Ozaki, Shuhei Yokoo. Large-scale Landmark Retrieval/Recognition under a Noisy and Diverse Dataset. 2019
[4] https://www.kaggle.com/c/landmark-retrieval-2019
[5] Mingxing Tan, Quoc V. Le. EfficientNet: Rethinking Model Scaling for Convolutional Neural Networks. ICML, 2019
[6] Hao Wang, Yitong Wang, Zheng Zhou, Xing Ji, Dihong Gong, Jingchao Zhou, Zhifeng Li, and Wei Liu∗. CosFace: Large Margin Cosine Loss for Deep Face Recognition. CVPR, 2018
[7] https://github.com/qubvel/efficientnet
[8] AdaCos : Adaptively Scaling Cosine Logits for Effectively Learning Deep Face Representations. CVPR, 2019
[9] https://colab.research.google.com